\documentclass{article}

\usepackage[utf8]{inputenc} 
\usepackage[T1]{fontenc}    
\usepackage[sc,osf]{mathpazo}
\usepackage[height=8.5in,width=6in,letterpaper]{geometry}
\usepackage[colorlinks=true,linkcolor=red,citecolor=blue]{hyperref}       
\usepackage{url}            
\usepackage{booktabs}       
\usepackage{amsfonts}       
\usepackage{nicefrac}       
\usepackage{microtype}      

\usepackage{epsfig}
\usepackage{graphicx}
\usepackage{amsmath}
\usepackage{amssymb}
\usepackage[numbers]{natbib}
\usepackage{bm}

\newcommand{\reals}[1]{\mathbb{R}^{#1}}
\newcommand{\enorm}[1]{\left\|{#1}\right\|}
\newcommand{\hilb}[1]{\left\|{#1}\right\|_{\mathcal{H}}}
\newcommand{\myexp}[1]{e^{\left({#1}\right)}}

\DeclareMathOperator*{\trace}{Tr}
\newcommand{\half}{\frac{1}{2}}

\renewcommand\cdots{...}

\newcommand{\suptensor}[1]{\mathfrak{S}^{d}}

\DeclareMathOperator*{\argmin}{arg\,min}
\newcommand{\fnorm}[1]{\left\|{#1}\right\|_F}

\newcommand{\comment}[1]{}

\newcommand{\ortho}{\mathcal{O}}
\newcommand{\grass}[1]{\mathcal{G}{#1}}

\DeclareMathOperator*{\subjectto}{\text{subject to}}
\DeclareMathOperator*{\sym}{sym}

\newcommand{\eye}[1]{\mathbf{I}_{#1}}
\newcommand{\mK}{\mathbf{K}}

\newcommand{\vx}{\mathbf{x}}
\newcommand{\mZ}{Z}
\newcommand{\vz}{\mathbf{z}}
\newcommand{\mX}{X}
\newcommand{\mU}{U}

\DeclareMathOperator*{\viol}{viol}
\newcommand{\kernel}{\mathbf{k}}
\newcommand{\mV}{V}
\newcommand{\mA}{A}
\newcommand{\mS}{S}
\newcommand{\mP}{P}
\newcommand{\mR}{R}
\newcommand{\mL}{L}
\newcommand{\mF}{F}
\usepackage{subfigure}

\makeatletter
\newlength\aftertitskip     \newlength\beforetitskip
\newlength\interauthorskip  \newlength\aftermaketitskip

\def\maketitle{\par
 \begingroup
   \def\thefootnote{\fnsymbol{footnote}}
   \def\@makefnmark{\hbox to 4pt{$^{\@thefnmark}$\hss}}
   \@maketitle \@thanks
 \endgroup
\setcounter{footnote}{0}
 \let\maketitle\relax \let\@maketitle\relax
 \gdef\@thanks{}\gdef\@author{}\gdef\@title{}\let\thanks\relax}

\def\@startauthor{\noindent \normalsize\bf}
\def\@endauthor{}
\def\@starteditor{\noindent \small {\bf Editor:~}}
\def\@endeditor{\normalsize}
\def\@maketitle{\vbox{\hsize\textwidth
 \linewidth\hsize \vskip \beforetitskip
 {\begin{center} \LARGE\@title \par \end{center}} \vskip \aftertitskip
 {\def\and{\unskip\enspace{\rm and}\enspace}%
  \def\addr{\small\it}%
  \def\email{\hfill\small\tt}%
  \def\name{\normalsize\bf}%
  \def\AND{\@endauthor\rm\hss \vskip \interauthorskip \@startauthor}
  \@startauthor \@author \@endauthor}
}}

\makeatother

\begin{document}

\title{Sequence Summarization Using Order-constrained Kernelized Feature Subspaces}

\author{\name Anoop Cherian
  \email{anoop.cherian@anu.edu.au}\\
  \addr{Australian Centre for Robotic Vision, Australian National University, Canberra}\\
  \name Suvrit Sra\email{suvrit@mit.edu}\\
  \addr{Massachusetts Institute of Technology, Cambridge, MA, USA}\\
  \name Richard Hartley \email{richard.hartley@anu.edu.au}\\
\addr{Australian Centre for Robotic Vision, Australian National University, Canberra}\\
}

\maketitle

\begin{abstract}
Representations that can compactly and effectively capture temporal evolution of semantic content are important to machine learning algorithms that operate on multi-variate time-series data. We investigate such representations motivated by the task of human action recognition. Here each data instance is encoded by a multivariate feature (such as via a deep CNN) where action dynamics are characterized by their variations in time. As these features are often non-linear, we propose a novel pooling method, \emph{kernelized rank pooling}, that represents a given sequence compactly as the pre-image of the parameters of a hyperplane in an RKHS, projections of data onto which captures their temporal order. We develop this idea further and show that such a pooling scheme can be cast as an order-constrained kernelized PCA objective; we then propose to use the parameters of a kernelized low-rank feature subspace as the representation of the sequences. We cast our formulation as an optimization problem on generalized Grassmann manifolds and then solve it efficiently using Riemannian optimization techniques. We present experiments on several action recognition datasets using diverse feature modalities and demonstrate state-of-the-art results. 
\end{abstract}


\section{Introduction}
\label{sec:intro}
Multivariate time-series data are ubiquitous in machine learning applications. A few notable examples include data from health monitoring devices, visual surveillance cameras, weather forecasting systems, and multimedia applications. Such data are often very high-dimensional and grow continually with time. As a result, obtaining compact representations that capture their intrinsic semantic content are of paramount importance to applications. 

In this paper, we propose compact representations for non-linear multivariate data arising in computer vision applications, by casting them in the concrete setup of action recognition in video sequences. 

The concrete setting we pursue is quite challenging. Although, rapid advancement of deep convolutional neural networks has led to significant breakthroughs in several computer vision tasks (e.g., object recognition, face recognition), action recognition continues to be significantly far from human-level performance. This gap is perhaps due to the spatio-temporal nature of the data involved due to which their size quickly outgrows processing capabilities of even the best hardware platforms. To tackle this, deep learning algorithms for video processing usually consider subsequences (a few frames) as input, extract features from such clips, and then aggregate these features into compact representations, which are then used to train a classifier for recognition.

In the popular two-stream CNN architecture for action recognition~\cite{simonyan2014two,feichtenhofer2016convolutional}, the final classifier scores are fused using a linear SVM. A similar strategy is followed by other more recent approaches such as the 3D convolutional network~\cite{tran2015learning} and temporal segment networks~\cite{wang2015action}. Given that an action is comprised of ordered variations of spatio-temporal features, any pooling scheme that discards such temporal variation may lead to sub-optimal performance. 

Consequently, various temporal pooling schemes have been devised. One recent scheme that showed promise is that \emph{rank pooling}~\cite{fernando2015modeling,Fernando:ICML2016}, in which the temporal action dynamics are summarized as the parameters of a line in the input space that preserves the frame order via linear projections. To estimate such a line, a rank-SVM~\cite{cao2007learning} based formulation is proposed, where the ranking constraints enforce the temporal order (see Section.~\ref{sec:background}). However, this formulation is limited on several fronts, notably (i) it assumes the data belongs to a linear vector space or are vectorial objects (and thus cannot handle structured objects such positive definite matrices, strings, trees, etc.), (ii) only linear ranking constraints are used, however non-linear projections may prove more fruitful, and (iii) data is assumed to evolve smoothly (or needs to be explicitly smoothed) as otherwise the pooled descriptor may fit to random noise. 

In this paper, we introduce~\emph{kernelized rank pooling} (KRP) that aggregates data features after mapping them to a (potentially) infinite dimensional reproducing kernel Hilbert space (RKHS) via a feature map. Our scheme learns hyperplanes in the feature space that encodes the temporal order of data via inner products; the pre-images of such hyperplanes in the input space are then used as action descriptors, which can then be used in a non-linear SVM for classification. This appeal to kernelization generalizes rank pooling to any form of data for which a Mercer kernel is available, and thus naturally takes care of challenges described above. We explore variants of this basic KRP in Section~\ref{sec:krp}.

A technical difficulty with KRP is its reliance on the computation of a pre-image of a point in feature space. However, given that the pre-images are finite-dimensional representatives of infinite-dimensional Hilbert space points, they may not be unique or may not even exist~\cite{mika1998kernel}. To this end, we propose an alternative kernelized pooling scheme based on feature subspaces (KRP-FS) in which instead of estimating a single hyperplane in the feature space, we estimate a low-rank kernelized subspace subject to the constraint that projections of the kernelized data points into this subspace should preserve temporal order. We propose to use the parameters of this low-rank kernelized subspace as the action descriptor. To estimate the descriptor, we propose a novel  order-constrained low-rank kernel approximation, with orthogonality constraints on the estimated descriptor. While, our formulation looks computationally expensive at first glance, we show that it allows efficient solutions if resorting to Riemannian optimization schemes on a generalized Grassmann manifold (Section~\ref{sec:efficient_optimization}). 

We present experiments on a variety of action recognition datasets, using different data modalities, such as CNN features from single RGB frames, optical flow sequences, trajectory features, pose features, etc. Our experiments clearly show the advantages of the proposed schemes achieving state-of-the-art results. 

Before proceeding, we summarize below the main contributions of this paper.
\begin{itemize}
\setlength{\itemsep}{1pt}
\item We introduce a novel~\emph{order-constrained kernel PCA} objective that learns action representations in a kernelized feature space. We believe our formulation may be of independent interest in other applications.
\item We introduce a new pooling scheme,~\emph{kernelized rank pooling} based on kernel pre-images that captures temporal action dynamics in an infinite-dimensional RKHS. 
\item We propose efficient Riemannian optimization schemes on the generalized Grassmann manifold for solving our formulations.
\item We show experiments on several datasets demonstrating state-of-the-art results.
\end{itemize}


\section{Related Work}
\label{sec:related_work}
Data mining and machine learning on time-series is a topic that has witnessed significant attention over the years and thus we limit our literature survey to recent methods proposed for the problem of activity recognition in video sequences. For classic approaches to time-series data analysis, we recommend surveys such as Keogh et al.~\cite{keogh2004segmenting}.

As alluded to earlier, recent methods for video based action recognition use features from the intermediate layers of a CNN, such features are then pooled into compact representations. The popular two-stream CNN model ~\cite{simonyan2014two} for action recognition has been extended using more powerful CNN architectures incorporating intermediate feature fusion in ~\cite{feichtenhofer2016convolutional,feichtenhofer2016spatiotemporal,tran2015learning}, however independent action predictions or features are still pooled independent of their temporal order during the final sequence classification. Wang et al.~\cite{Wang2016} enforces a grammar on the two-stream model via temporal segmentation, however this grammar is designed manually. Another popular approach for action recognition has been to use recurrent networks in the form of RNNs and LSTMS~\cite{yue2015beyond,donahue2014long}. However, training such models is often difficult~\cite{pascanu2013difficulty} or would need enormous datasets.

Amongst recently proposed temporal pooling schemes, rank pooling~\cite{fernando2015modeling} has witnessed a lot of attention due to its simplicity and effectiveness. There have been extension of this scheme in a discriminative setting~\cite{fernando2016discriminative,bilen2016dynamic,Fernando:ICML2016, dynamic_flow}, however all these variants use the basic rank-SVM formulation and is limited in their representational capacity as alluded to in the last section. Recently, in Cherian et al.~\cite{cherian_grp}, the basic rank pooling is extended to use the parameters of a feature subspace, however their formulation also assumes data embedded in the Euclidean space. In contrast, we generalize rank pooling to any form of data that has a valid similarity kernel. We further extend this idea by formulating an order-constrained kernel PCA to learn kernelized feature subspaces as data representations. To the best of our knowledge, both these ideas have not been proposed previously. 

We note that kernels have been used to describe actions earlier. For example, Cavazza et al.~\cite{cavazza2016kercov} and Quang et al.~\cite{quang2016approximate} propose kernels capturing spatio-temporal variations for action recognition, where the geometry of the SPD manifold is used for classification. Koniusz et al.~\cite{koniusz2016tensor} uses pose sequences embedded in an RKHS, however the resultant kernel is linearized and embedded in the Euclidean space. Vemulapalli et al.~\cite{vemulapalli2014human} uses SE(3) geometry to classify pose sequences. Tseng~\cite{Tseng2012} proposes to learn a low-rank subspace where the action dynamics are linear. Subspace representations have also been investigated in~\cite{turaga2011statistical,harandi2013kernel,OHara2012}, and the final representations are classified using Grassmannian kernels.  However, we differ from all these schemes in that our subspace is learned using temporal order constraints, and our final descriptor is an element of the RKHS, offering greater flexibility and representational power in capturing non-linear action dynamics. We also note that there have been extensions of kernel PCA for computing pre-images, such as for denoising~\cite{mika1998kernel,kwok2004pre}, voice recognition~\cite{kwok2004eigenvoice}, etc., but are different from ours in methodology and application.

\section{Preliminaries}
\label{sec:background}
In this section, we setup the notation for the rest of the paper and review some prior formulations for pooling multivariate time series for action recognition. Let $\mX=\left[\vx_1, \vx_2,\cdots, \vx_n\right]$ be features from $n$ consecutive frames of a video sequence, where we assume each $\vx_i\in\reals{d}$. 

Rank pooling~\cite{fernando2015modeling} is a scheme to compactly represent a sequence of frames into a single feature that summarizes the sequence dynamics. Typically, rank pooling solves the following objective:
\begin{equation}
\\arg\min_{\vz\in\reals{d}} \half\enorm{\vz}^2 + \lambda\sum_{i<j} \max(0, \eta + \vz^T \vx_i -\vz^T \vx_j),
\label{eq:rank-basic}
\end{equation}
where $\eta > 0$ is a threshold enforcing the temporal order and $\lambda$ is a regularization constant. Note that, the formulation in~\eqref{eq:rank-basic} is the standard Ranking-SVM formulation~\cite{cao2007learning} and hence the name. The minimizing  vector $\vz$ (which captures the parameters of a line in the input space) is then used as the pooled action descriptor for $\mX$ in a subsequent classifier. The rank pooling formulation in~\cite{fernando2015modeling} encodes the temporal order as increasing intercept of input features when projected on to this line. 

The objective in~\eqref{eq:rank-basic} only considers preservation of the temporal order, while, the minimizing $\vz$ may not be related to 
input data at a semantic level (as there are no constraints enforcing this). It may be beneficial for $\vz$ to capture some discriminative properties of the data (such as human pose, objects in the scene, etc.), that may help subsequent classifier. To account for these shortcomings, Cherian et al.~\cite{cherian_grp}, extended rank pooling to use the parameters of a subspace as the representation for the input features with better empirical performance. Specifically,~\cite{cherian_grp} solves the following problem.
\begin{equation}
\label{eq:low-rank}
\min_{\mU\in\grass(p,d)} \half\sum_{i=1}^n \enorm{\vx_i - \mU\mU^T\vx_i}^2 + \sum_{i<j}\max(0, \eta + \enorm{\mU^T \vx_i}^2 - \enorm{\mU^T \vx_j}^2),
\end{equation}
where instead of a single $\vz$ as in~\eqref{eq:rank-basic}, we learn a subspace $\mU$ (belonging to a $p$-dimensional Grassmann manifold embedded in $\reals{d}$), such that this $\mU$ provides a low-rank approximation to the data, as well as, projection of the data points onto this subspace will preserve their temporal order in terms of their distance from the origin.

However, both the above schemes have limitations; they assume input data is vectorial; as a result, they cannot be directly applied to temporal sequences of matrices, tensors, etc. Further, the input features are assumed to belong to a vector space, which may be severely limiting when working with features from an inherently non-linear space. To circumvent these issues, in this paper, we explore kernelized rank pooling schemes that generalize rank pooling to arbitrary data objects, for which a valid Mercer kernel can be computed. In the sequel, we assume an RBF kernel for the feature map, defined for $\vx,\vz\in\reals{d}$ as:
\begin{equation}
\kernel(\vx, \vz) = \exp\left\{-\frac{\enorm{\vx-\vz}^2}{2\sigma^2}\right\},
\label{eq:rbf}
\end{equation}
for a bandwidth $\sigma$. We use $\mK$ to denote the $n\times n$ RBF kernel matrix constructed on all frames in $\mX$, i.e., the $ij$-th element $\mK_{ij} = \kernel(\vx_i, \vx_j)$, where $\vx_i,\vx_j\in\mX$.
\section{Our Approach}
\label{sec:krp}
Given a sequence of temporally-ordered (potentially non-linear) features~$\mX$, our main idea is to use the kernel trick to map the features to a (plausibly) infinite-dimensional RKHS, where the data is linear. We propose to learn a hyperplane in the feature space, projections of the data to which will preserve the temporal order. We formalize this idea below and explore variants.

\subsection{Kernelized Rank Pooling}
\label{sec:bkrp}
Suppose, for a data point $\vx\in\reals{d}$, let $\Phi(\vx)$ be its embedding in an RKHS. Then, a straightforward way to extend~\eqref{eq:rank-basic} is to use a direction $\Phi(\vz)$ in the feature space, projections of $\Phi(\vx)$ onto this line will preserve the temporal order. However, given that we need to retrieve $\vz$ in the input space, to be used as the pooled descriptor ($\Phi(\vz)$ could potentially be infinite dimensional), we propose to compute the pre-image $\vz$ of $\Phi(\vz)$, which is then used as the action descriptor in a subsequent non-linear action classifier. Mathematically, our~\emph{basic kernelized rank pooling} (BKRP) formulation is as follows:
\begin{align}
\arg\min_{\vz\in\reals{d}}\quad \text{BKRP}(\vz) &:= \half\enorm{\vz}^2 + \lambda\sum_{i<j} \max(0, \eta + \langle \Phi(\vx_i), \Phi(\vz)\rangle - \langle \Phi(\vx_j), \Phi(\vz)\rangle \\
&= \half\enorm{\vz}^2 + \lambda \sum_{i<j} \max(0, \eta + \kernel(\vx_i, \vz) - \kernel(\vx_j, \vz)).
\label{eq:bkrp}
\end{align}

As alluded to earlier, a technical issue with~\eqref{eq:bkrp} (and so with~\eqref{eq:rank-basic}) is that the optimal direction $\vz$ might ignore any important properties of the original data $\mX$ (it could be some arbitrary direction that preserves the temporal order alone). To ascertain that the pre-image $\vz$ is similar to $\vx\in\mX$, we rewrite an improved~\eqref{eq:bkrp} (after incorporating slack variables $\xi_{ij}$) as:
\begin{align}
\argmin_{\vz\in\reals{d},\xi\geq 0} \text{IBKRP}(\vz)\!&:=\half\sum_{i=1}^n\enorm{\vx_i-\vz}^2 + C\sum_{i,j=1}^n\!\!\xi_{ij}\!+\lambda\sum_{i<j} \max(0, \eta-\xi_{ij} + \kernel\left(\vx_i, \vz) - \kernel(\vx_j, \vz)\right),
\end{align}
where the first component ensures that the computed pre-image is not far from the input data.\footnote{When $\vx$ is not an object in the Euclidean space, we assume $\enorm{ . }$ to define some suitable distance on the data.} The variables $\xi$ represent non-negative slacks and $C$ is a positive constant. 

The above formulation assumes a pre-image always exist, which may not be the case, or may not be unique even if it exists~\cite{mika1998kernel}. In the following, we assume that useful data maps to a low-rank $p$-dimensional subspace of the feature space, and in this subspace, the temporal order of data points is preserved (in some way). We propose to use the parameters of this subspace as the representation of data. In contrast to a single $\vz$ that captures action dynamics, a subspace offers significantly more representational capacity and as will be clear in the next section, thanks to the representation theorem, we can avoid the need to compute a pre-image, instead can use the parameters of the RKHS subspace directly as our action descriptor using a Grassmannian kernel. 

\subsection{Kernelized Rank Pooling Using Feature Subspaces}
Before describing our method in detail, we will need some notation. While, we assume $\Phi(\vx)$ is a feature map to an infinite dimensional RKHS, we assume that useful data in fact belongs to a  finite $p$-dimensional subspace embedded in this RKHS. We use $\Omega_p(\Phi(\vx))$ to denote this embedding of the input data point $\vx\in\reals{d}$. If $\Omega_p = \mV$ defines this low-rank subspace, then by the Representer theorem~\cite{wahba1990spline,scholkopf2001generalized}, $\Omega_p(\Phi(\vx)) = \mV\beta$ defines the reconstruction of $\Phi(\vx)$ in $\mV$ where $\beta$ captures the projection coefficients, i.e., $\beta_i=\mV_j^T\Phi(\vx)$, $\mV_j$ is the $j$-th basis of $\mV$. As is well-known~\cite{mika1998kernel}, $\mV$ can be written as a linear combination of the mapped data points, i.e., for a coefficient matrix $\mA$, we can write $\mV = \Phi(\mX)\mA$, where $\Phi(\mX)$ defines the feature map for $\mX$ and $\mA$ is an $n\times p$ matrix. Then, using these substitutions, $\Omega_p(\Phi(\vx))$ can be written as:
\begin{equation}
\Omega_p(\Phi(\vx)) = \Phi(\mX)\mA\mA^T\kernel(\mX, \vx),
\label{eq:omega}
\end{equation}
where (with a slight abuse of notation) we have defined $\kernel(\mX, \vx)$ to denote the $n\times 1$ vector whose $j$-th dimension is given by $\kernel(\vx_j,\vx)$.

Using these notation, we propose a novel kernelized feature subspace learning (KRP-FS) formulation below, with ordering constraints: 
\begin{align}
\label{eq:krpfs-obj}
&\arg\min_{\mA\in\reals{n\times p}} \quad \mF(\mA) := \half\sum_{i=1}^n   \hilb{\Phi(\vx_i) - \Omega_p(\Phi(\vx_i))}^2 &\\
&\subjectto\qquad \hilb{\Omega_p(\Phi(\vx_i))}^2 + \eta \leq \hilb{\Omega_p(\Phi(\vx_j))}^2,\quad\forall i<j,
\label{eq:krpfs-cons}
\end{align}
where~\eqref{eq:krpfs-obj} learns a $p$-dimensional feature subspace of the RKHS in which most of the data variability is contained, while~\eqref{eq:krpfs-cons} enforces the temporal order of data in this subspace, as measured by the length (using a Hilbertian norm $\hilb{.}$) of the projection of the data points $\vx$ onto this subspace. Our main idea is to use the parameters of this kernelized feature subspace $\Omega_p$ as the pooled descriptor for our input sequence, for subsequent sequence classification. To ensure that such descriptors from different sequences are comparable, we need to ensure that $\Omega_p$ is normalized, i.e., has orthogonal columns in the features space, that is, $\Omega_p^T \Omega_p= \eye{p}$. Substituting the definition of $\Omega_p=\Phi(\mX)\mA$ from~\eqref{eq:omega}, this condition boils down to:
\begin{equation}
\Omega_p^T\Omega_p = \mA^T \langle \Phi(\mX), \Phi(\mX)\rangle \mA = \mA^T\mK\mA = \eye{p},
\label{eq:gengrass}
\end{equation}
where $\mK$ is the kernel constructed on $\mX$ and is symmetric positive definite (SPD). Incorporating these conditions and including slack variables, we rewrite~\eqref{eq:krpfs-obj} as:
\begin{align}
\label{eq:krpfs-obj-1}
&\argmin_{\substack{\mA\in\reals{n\times p}| \mA^T\mK\mA = \eye{p}\\ \xi\geq 0}} \mF(\mA) := \half\sum_{i=1}^n \hilb{\Phi(\vx_i) - \Omega_p(\Phi(\vx_i))}^2  + C\sum_{i<j}^n \xi_{ij} &\\
&\subjectto\qquad \hilb{\Omega_p(\Phi(\vx_i))}^2 + \eta -\xi_{ij} \leq \hilb{\Omega_p(\Phi(\vx_j))}^2,\quad\forall i<j.
\label{eq:krpfs-cons-1}
\end{align}

It may be noted that our objective~\eqref{eq:krpfs-obj-1} essentially depicts kernel principal components analysis (KPCA)~\cite{scholkopf1997kernel}, albeit the constraints make estimation of the low-rank subspace different, demanding sophisticated optimization techniques for efficient solution. We address this concern below, by making some key observations regarding our objective.

\subsection{Efficient Optimization}
\label{sec:efficient_optimization}
After substituting for the definitions of $\mK$ and $\Omega_p$, the formulation in~\eqref{eq:krpfs-obj-1} is rewritten using hinge-loss as:
\begin{align}
&\argmin_{\substack{\mA\in\reals{n\times p}| \mA^T\mK\mA = \eye{p}\\ \xi\geq 0}}\hspace*{-0.7cm}\mF(\mA) := \half\sum_{i=1}^n -2\kernel(\mX,\vx_i)^T\mA\mA^T\kernel(\mX,\vx_i) + \kernel(\mX,\vx_i)^T\mA\mA^T \mK \mA\mA^T\kernel(\mX, \vx_i)+C\sum_{i<j}\xi_{ij}\nonumber\\ 
& + \lambda \sum_{i<j} \max(0, \kernel(\mX, \vx_i)^T\mA\mA^T\mK\mA\mA^T\kernel(\mX,\vx_i) + \eta -\xi_{ij} - \kernel(\mX,\vx_j)^T\mA\mA^T\mK\mA\mA^T\kernel(\mX,\vx_j),
\label{eq:krpfs-hinge-loss}
\end{align}
As is clear, the variable $\mA$ appears as $\mA\mA^T$ through out and thus our objective is invariant to any right rotations by an element of the $p$-dimensional orthogonal group $\ortho(p)$, ie., for $\mR\in\ortho(p), \mF(\mA) = \mF(\mA\mR)$. This, together with the condition in~\eqref{eq:gengrass} suggests that the optimization on $\mA$ as defined in~\eqref{eq:krpfs-hinge-loss} can be solved over the so called~\emph{generalized Grassmann manifold}~\cite{edelman1998geometry}[Section 4.5] using Riemannian optimization techniques. 

We use a Riemannian conjugate gradient (RCG) algorithm on this manifold for solving our objective. A key component for this algorithm to proceed is the expression for the Riemannian gradient of the objective $\mathrm{grad}_{A} \mF(\mA)$, which for a generalized Grassmannian can be obtained from the Euclidean gradient $\nabla_{\mA}\mF(\mA)$ as:
\begin{equation}
\mathrm{grad}_\mA\mF(\mA) = \mK^{-1}\nabla_{\mA}\mF(\mA) - \mA\sym(A^T\nabla_{\mA} \mF(\mA)),
\end{equation}
where $\sym(\mL) = \frac{L+L^T}{2}$ is the symmetrization operator~\cite{mishra2016riemannian}[Section 4]. The Euclidean gradient for~\eqref{eq:krpfs-hinge-loss} is as follows: let $\mS_1=\mK\mK\mA$, $\mS_2=\mK\mA\mA^T$, and $\mS_3=\mA^T\mK\mA$, then
\begin{align}
\nabla_{\mA}\mF(\mA) &= \mS_1\left(\mS_3 - 2\eye{p}\right) + \mS_2\mS_1 + \lambda\left(\mK_{12}\mA\mS_3 + \mS_2\mK_{12}\mA\right),
\label{eq:euc_grad}
\end{align}
where $\mK_{12}= \mK_1\mK_1^T - \mK_2\mK_2^T$, $\mK_1,\mK_2$ are the kernels capturing the order violations in~\eqref{eq:krpfs-hinge-loss} for which the hinge-loss is non-zero -- $\mK_1$ collecting the sum of all violations for $\vx_i$ and $\mK_2$ the same for $\vx_j$. If further scalability is desired, one can also invoke stochastic Riemannian solvers such as Riemannian-SVRG~\cite{zhang2016riemannian,kasai2016riemannian} instead of RCG. These methods extend the variance reduced stochastic gradient methods to Riemannian manifolds, and may help scale the optimization to larger problems.

\subsection{Action Classification}
Once we find $\mA$ per video sequence by solving~\eqref{eq:krpfs-hinge-loss}, we use $\Omega = \Phi(\mX)\mA$ (note that we omit the subscript $p$ from $\Omega_p$ as it is by now understood) as the action descriptor. However, as $\Omega$ is semi-infinite, it cannot be directly computed, and thus we need to resort to the kernel trick again for measuring the similarity between two encoded sequences. Given that $\Omega\in\grass(p)$ belonging to a generalized Grassmann manifold, we can use any Grassmannian kernel for computing this similarity. Among several such kernels reviewed in~\cite{harandi2014expanding}, we found the exponential projection metric kernel to be empirically beneficial. For two sequences $\mX_1\in\reals{d\times n_1},\mX_2\in\reals{d\times n_2}$, their subspace parameters $\mA_1,\mA_2$ and their respective KRP-FS descriptors $\Omega_1,\Omega_2\in\grass(p)$, the exponential projection metric kernel is defined as:
\begin{equation}
\mathbb{K}_{\mathcal{G}}(\Omega_1, \Omega_2) = \exp\left(\nu \fnorm{\Omega_1^T\Omega_2}^2\right), \text{for } \nu>0.
\end{equation}
Substituting for $\Omega$'s, we have the following kernel for action classification, whose $ij$-th entry is given by:
\begin{align}
\mathbb{K}_{\mathcal{G}}^{ij}(\Omega_1,\Omega_2) &= \exp\left(\nu\fnorm{\mA_i \mK \mA_j}^2\right),
\end{align}
where $\mK\in\reals{n_1\times n_2}$ is an (RBF) kernel capturing the similarity between sequences and whose $rs$-th element is given by $\mK_{rs}=\kernel(\vx^1_r,\vx^2_s)$, $\vx^1_r\in \mX_1$ and $\vx^2_s\in\mX_2$ (using notation defined in~\eqref{eq:rbf}).

\section{Computational Complexity}
Evaluating the objective in~\eqref{eq:krpfs-hinge-loss} takes $\mathcal{O}(n^2p)$ operations and computing the Euclidean gradient in~\eqref{eq:euc_grad} needs $\mathcal{O}(n^3+n^2p)$ computations for each iteration of the solver. While, this may look more expensive than the basic ranking pooling formulation, note that here we use kernel matrices, which for action recognition datasets, are much smaller in comparison to very high dimensional (CNN) features used for frame encoding, and note that basic rank pooling operates on such high-dimensional features.

\section{Experiments}
\label{sec:expts}
In this section, we provide experiments on several action recognition datasets where action features are represented in diverse ways. Our goal is to show the effectiveness of the proposed pooling schemes, irrespective of the feature type used. Towards this end, we experiment on (i) the JHMDB dataset, where frames are encoded using the features from the intermediate layers (fc6) of a VGG two-stream CNN model (details below), (ii) UTKinect actions dataset, which consist of 3D skeleton sequences corresponding to human actions, and (iii) MPII cooking activities dataset, for which we use extracted dense trajectory features for action modeling. For all these datasets, we compare our performance to basic rank pooling (RP) scheme, as well as to state-of-the-art methods, such as generalized rank pooling (GRP)~\cite{cherian_grp}. We used the publicly available code for rank pooling~\cite{fernando2015modeling} without any modification.\comment{, while for GRP, we implemented the algorithm ourselves} Our implementations are in Matlab. Below, we first describe our datasets and their pre-processing, following which we furnish our experimental results.

\subsection{Datasets and Feature Representations}
\paragraph*{\small{JHMDB Dataset~\cite{jhuang2013towards}}}consists of 960 video sequences and 21 actions, each with 10-40 frames.
We used the split-1 of this dataset for our evaluation consisting of 660 sequences for training/validation and the rest for testing. As mentioned above, we use a two-stream VGG CNN architecture for frame encoding, that has two separate VGG-16 networks, one taking single RGB frames and the other one using a small stack of 10 consecutive optical flow images. To fine-tune the network on our dataset, we started with a similar model for the UCF-101 dataset provided as part of~\cite{feichtenhofer2016convolutional}. The two streams are fine-tuned separately (at a learning rate of $10^{-4}$, for 40K and 20K iterations respectively for flow and RGB, see~\cite{cherian2017second} for details of training). Later, we extract features from the fc6 layers of the streams for every video frame (and flow stack), and use as frame-level features in our KRP scheme. 

\paragraph*{\small{UTKinect Actions~\cite{xia2012view}}}is a dataset for human action recognition from 3D skeleton sequences. UTKinect actions consist sequences with 74 frames, 10 actions and performed by 2 subjects. We use the SE(3) encoding for the poses and use the evaluation criteria as described in~\cite{vemulapalli2014human}. 

\paragraph*{\small{MPII Cooking Activities Dataset~\cite{rohrbach2012database}}}consists of cooking actions of 14 individuals. The dataset has 5609 video sequences and 65 different actions. We use the dense trajectory features available as part of the dataset for our experiments. These trajectories are encoded using a bag-of-words model using 4000 words. We report the mean average precision over 7 splits of the dataset as is the standard practice.

\subsection{Parameter Estimation}
\begin{figure}[htbp]
\subfigure[]{\label{fig:1}\includegraphics[width=7cm, trim=0 7cm 0 8cm]{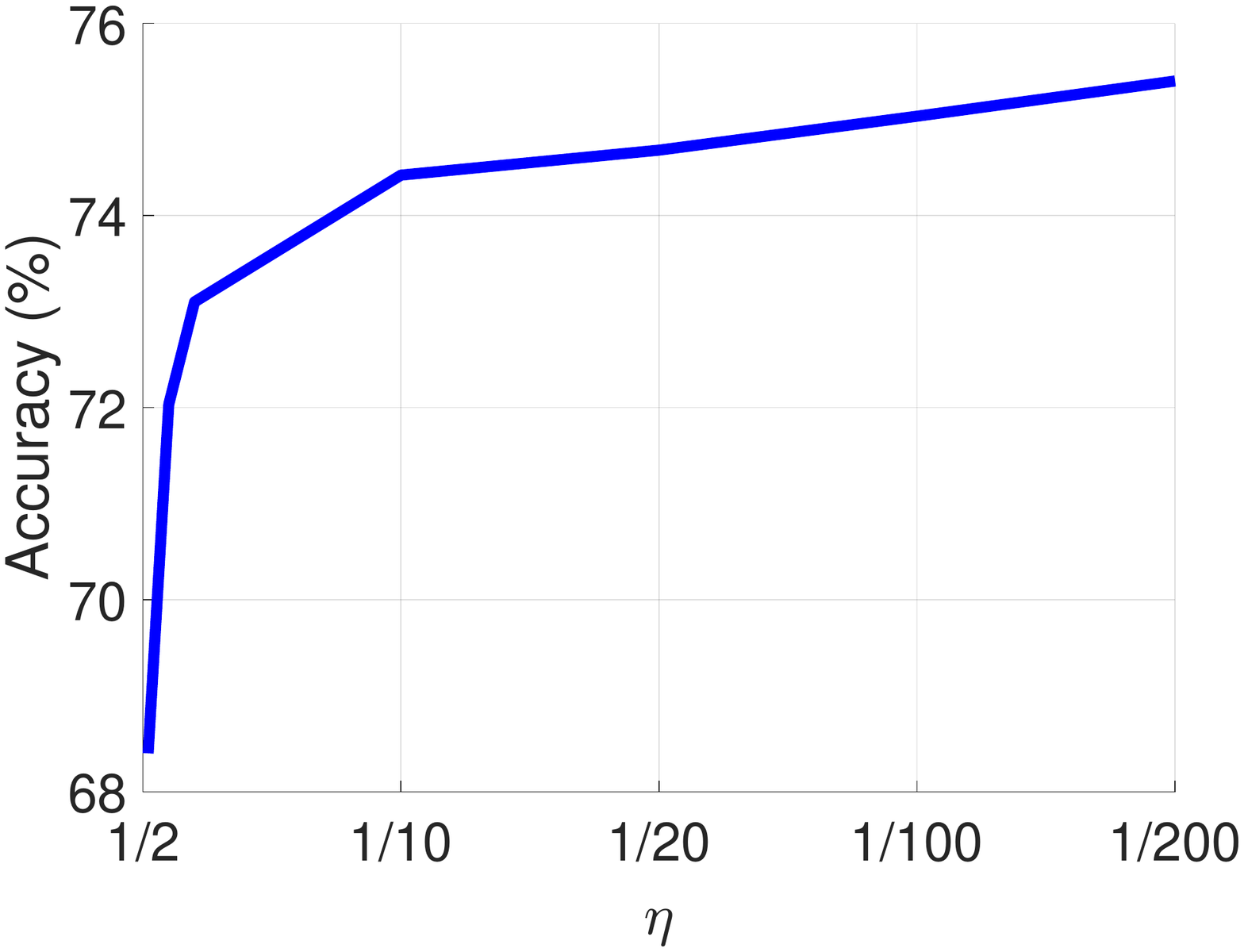}}
\subfigure[]{\label{fig:2}\includegraphics[width=7cm, trim=0 7cm 0 8cm]{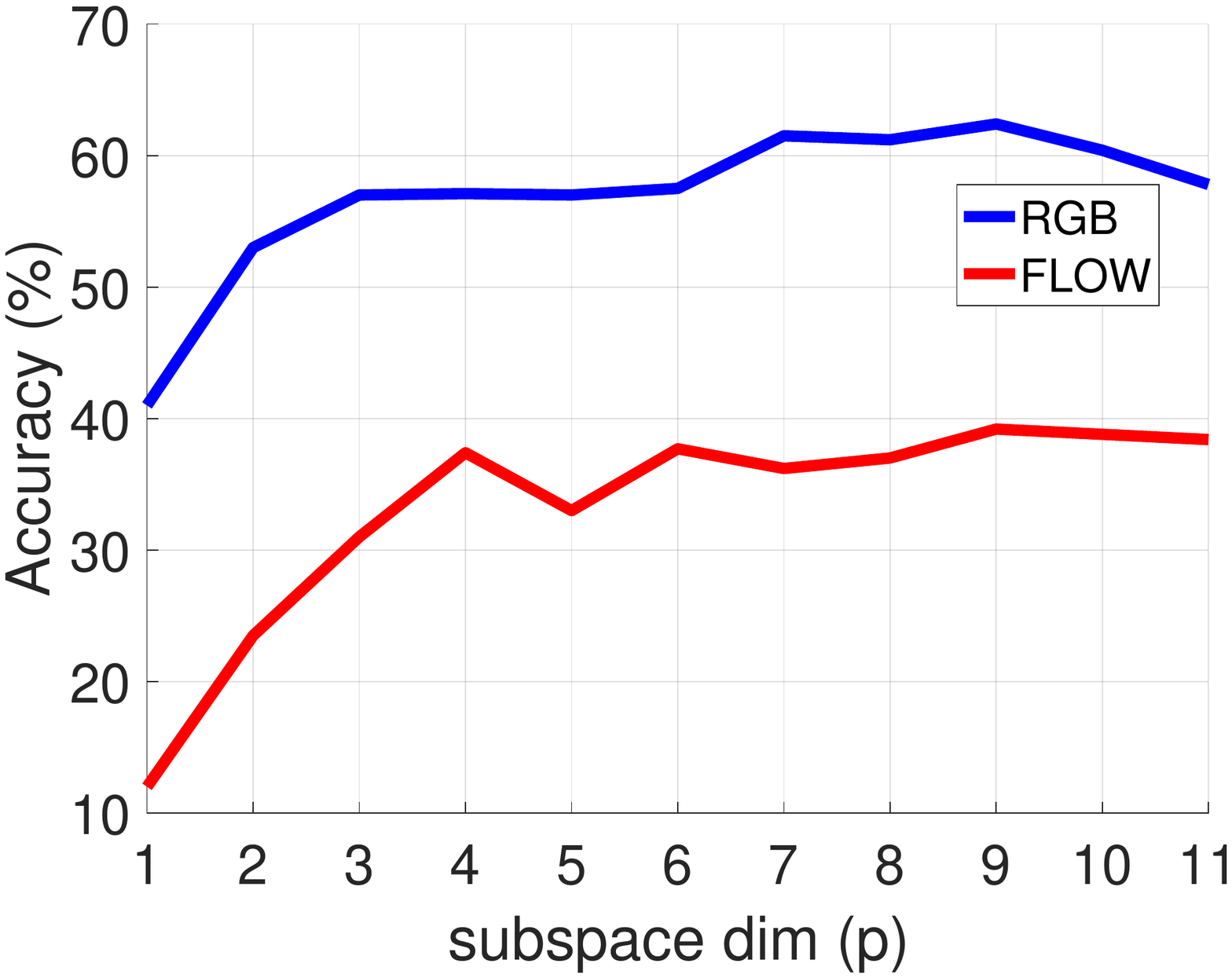}}
\caption{Left: Classification accuracy for decreasing ranking threshold ($\eta$) using IBKRP without slack variables. Right: accuracy of KRP-FS for increasing subspace dimensionality ($p$).}
\label{fig:parameter-plots}
\end{figure}

Below, we investigate the influence of two critical parameters in our setup, namely, the ranking threshold $\eta$ and the number of subspaces $p$ in KRP-FS. For the former experiment, we do not use slack parameters, and use our improved KRP objective (IBKRP) described in~\eqref{sec:bkrp}. We plot the results on the JHMDB dataset in Figure~\ref{fig:1}. As is clear from the plot, increasing the ranking threshold decreases the accuracy, which is unsurprising given that the lack of slack variables will force the algorithm to look at pre-images that are unrelated to data to satisfy the ranking constraints and thus leads to sub-optimal results. However, below about 0.01, the performance saturates. Note that the accuracy of average pooling on this dataset is about 71\% (Table~\ref{tab:jhmdb-results}), clearly showing that our ranking formulation is useful.

In Figure~\ref{fig:2}, we plot the classification accuracy of KRP-FS against increasing subspace dimensionality ($p$) for RGB and optical flow streams. As is clear from the plot, increasing $p$ helps improve accuracy, however beyond a certain value, the accuracy starts dropping, perhaps because of including noisy subspaces. Interestingly, both RGB and FLOW streams show a similar trend. In the sequel, we use $\eta=0.01$ and $p$ via cross-validation. As for the kernel bandwidth $\sigma$, we set it as the standard deviation of features in the respective sequence, which seemed to work well in practice. For optimization, we used the Manopt solver~\cite{boumal2014manopt}, and ran 100 iterations of conjugate gradient for all our different formulations. Note that the rank pooling formulation in~\cite{fernando2015modeling} need to use running means to smooth data before applying the ranking scheme. However, we found that this step is unnecessary in our setup, as perhaps the RBF kernel already looks for smooth data trajectories.

\subsection{Results}
In Tables~\ref{tab:jhmdb-results} and~\ref{tab:utkinect-results}, we show the evaluation of various ranking based pooling schemes alongside recent state-of-the-art results. As is clear from the tables, all our pooling schemes significantly improve the performance of linear rank pooling~\cite{fernando2015modeling} and the recent generalized rank pooling (GRP)~\cite{cherian_grp}. As expected, IBKRP is often better than BKRP (by about 3\%  on JHMDB and MPII, and nearly 8\% on UT Kinect actions). We also find that KRP-FS performs the best most often, with about 7\% better on the MPII cooking activities against GRP and UT Kinect actions. However, its performance seems inferior on the JHMDB dataset, which we believe is perhaps because of lack of sufficient number of frames in this dataset; JHMDB dataset has about 10-40 frames in each sequence, which may not be adequate for learning a useful subspace.

\begin{table}[htbp]
	\centering
	\begin{tabular}{|c|c|c|c|}
        \hline
		JHMDB Dataset &  FLOW  & RGB & FLOW+RGB \\				
		\hline
		Avg. ~\cite{simonyan2014two}         & 63.8 &  47.8 & 71.2  \\
        RP~\cite{Fernando2016} & 41.1 &  47.3 & 56.0  \\
        GRP~\cite{cherian_grp}     & 64.2 &  42.5 & 70.8 \\
        BKRP   (ours)               & 65.8 &  49.3 & 73.4 \\
        IBKRP  (ours)               & 68.2 &  49.0 & \textbf{76.2}\\
        KRP-FS (ours)              & 62.5 &  41.5 & 70.1\\
        \hline
	\end{tabular}
	\caption{Classification accuracy on the JHMDB dataset. In the second and third columns, we show the results of pooling the RGB and FLOW streams separately. Last column shows the results of combining the two pooled streams using an SVM.}           
    \label{tab:jhmdb-results}
\end{table}

\begin{table}[htbp]
	\centering
    \begin{tabular}{|c|c|}
       \hline
		MPII Dataset &  Trajectory \\				
		\hline
		Avg. Pooling & 42.1\\
        RP~\cite{fernando2015modeling} & 45.3\\
        GRP~\cite{cherian_grp} & 46.1\\
        BKRP (ours)  &  46.5\\
        IBKRP (ours) &  49.5\\
        KRP FS (ours) & \textbf{53.0}\\
        \hline
	\end{tabular}
	\begin{tabular}{|c|c|}
        \hline
	    UT Kinect &  Trajectory \\				
		\hline		
        Rank Pooling (linear) & 75.5\\        
        BKRP  &  84.8\\
        IBKRP &  92.1\\
        KRP FS (1 subspace)  & 94.1 \\      
        KRP FS (15 subspaces) & \textbf{99.0}  \\
        \hline
        SE(3) representations~\cite{vemulapalli2014human} & 97.1\\ 
        \hline
	\end{tabular}
	\caption{Left: Mean average precision on MPII cooking activities dataset (using trajectory features). Right: classification accuracy on the UT-Kinect actions dataset (using 3D pose skeleton sequences).}
      \label{tab:utkinect-results}
\end{table}


\section{Conclusions}
\label{sec:conclude}
In this paper, we proposed a novel kernelized ranking formulation for sequence summarization, where the pre-image of a temporal order preserving hyperplane in the RKHS was used as the data descriptor. We generalized this formulation, by introducing an order-constrained kernelized PCA objective, which may be useful beyond the applications proposed in this paper. We provided efficient Riemannian optimization algorithms for solving our formulation. Experiments were provided on various action recognition datasets encoding sequences using diverse feature representations. Our results clearly demonstrate the usefulness of our schemes irrespective of the data type, outperforming similar pooling schemes.

\bibliographystyle{plainnat}
\setlength{\bibsep}{3pt}
\bibliography{krp_bib}

\end{document}